\documentclass[10pt,article]{IEEEtran}
\usepackage{cite}
\usepackage{amsmath,amssymb,amsfonts}
\usepackage{graphicx}
\usepackage{textcomp}
\usepackage{xcolor}
\usepackage{tablefootnote}
\usepackage{url}
\usepackage{comment}
\def\BibTeX{{\rm B\kern-.05em{\sc i\kern-.025em b}\kern-.08em
T\kern-.1667em\lower.7ex\hbox{E}\kern-.125emX}}

\usepackage{acronym}
\acrodef{cnn}[CNN]{convolutional neural network}
\acrodef{dnn}[DNN]{deep neural network}
\acrodef{hmm}[HMM]{hidden Markov model}
\acrodef{pbn}[PBN]{projected belief network}
\acrodef{pbn-da}[PBN-DA]{discriminative alignment of \acl{pbn}}
\acrodef{pdf}[PDF]{probability density function}

\newcommand{\defined}{\stackrel{\mbox{\tiny$\Delta$}}{=}}

\newcommand{\bitem}{\begin{itemize}}

\newcommand{\eitem}{\end{itemize}}
\newcommand{\benum}{\begin{enumerate}}
\newcommand{\eenum}{\end{enumerate}}
\newcommand{\bdm}{\begin{displaymath}}

\newcommand{\edm}{\end{displaymath}}
\newcommand{\beq}{\begin{equation}}
\newcommand{\bea}{\begin{eqnarray}}
\newcommand{\eea}{\end{eqnarray}}

\newcommand{\barray}{\begin{displaymath} \begin{array}{rcl}}
\newcommand{\earray}{\end{array}\end{displaymath}}
\newcommand{\eeq}{\end{equation}}

\newcommand{\balpha}{\mbox{\boldmath $\alpha$}}

\newcommand{\bfu}{{\bf u}}
\newcommand{\bfx}{{\bf x}}

\newcommand{\bfh}{{\bf h}}

\newcommand{\bfz}{{\bf z}}

\begin{document}

\title{On Maximum Entropy Linear Feature Inversion}
\author{\IEEEauthorblockN{Paul M Baggenstoss}
\IEEEauthorblockA{\textit{Fraunhofer FKIE} \\
Fraunhoferstraße 20, 53343 Wachtberg, Germany\\
p.m.baggenstoss@ieee.org}

\thanks{This work was supported jointly by the Office of Naval Research Global and the Defense Advanced Research Projects Agency under Research Grant - N62909-21-1-2024}
}
\maketitle


\begin{abstract}
We revisit the classical problem of 
inverting dimension-reducing linear mappings
using the maximum entropy (MaxEnt) criterion.
In the literature, solutions are problem-dependent,
inconsistent, and use different entropy measures.
We propose a new unified approach that not only specializes to the existing 
approaches, but offers solutions to new cases, 
such as when data values are constrained to $[0,\; 1]$,
which has new applications in machine learning.
%Experiments confirm equivalence to classical 
%methods in one case, and superiority over 
%classical methods in another case.
\end{abstract}

\section{Introduction}

\subsection{Problem Statement}
Let $\bfx\in{\mathbb X} \subset {\mathbb R}^N$ be an unknown
input data sample. Let
%As an aside, the unbounded data case $[-\infty, \; \infty]$ is uninteresting since the least-squares
%solution is available, and meets myriad optimality criteria.
%through the output of 
\beq
\bfz = {\bf W}^T \bfx, 
\label{zdef}
\eeq
where ${\bf W}$ is a full-rank $N\times M$ matrix and $M<N$.
Given $\bfz$, we wish to reconstruct $\bfx$ by selecting
a member of the set
\beq
{\cal M}(\bfz) = \{ \bfx : \; {\bf W}^T \bfx = \bfz\}, \; \bfx \in {\mathbb X},
\label{manif}
\eeq
the set of all samples $\bfx\in {\mathbb X}$ 
that reproduce the $\bfz$.
%, in this case 
%the intersection of ${\mathbb X}$ with a linear subspace.
%Methods exist that do not restrict $\bfx$ to ${\cal M}(\bfz)$,
%but use an error term that allows data that is not feature-reproducing
%\cite{Zhuang87,Zhou,Wei87}, however we will assume that a solution exists.  
%Methods differ in how $\bfx$ is selected from ${\cal M}(\bfz)$,
%but all published approaches that we know about
%select the member of ${\cal M}(\bfz)$
%that satisfies some regularity condition such as
%minimum energy, maximum smoothness, or maximum entropy 
%\cite{Boucheron08,milner2002speech, Nadeu1990Flatness,Csiszar1991}.
Since ${\cal M}(\bfz)$ has an infinity of members, 
we must select a member based on some regularization criterion.
The leading criterion is maximum entropy (MaxEnt),
which is justified base on first principles 
\cite{Jaynes82,Khinchin,Csiszar1991,Nadeu1990Flatness,Reitsch77},
and can be seen as a general flatness measure \cite{Nadeu1990Flatness}.
Below, we revisit this classical problem, unify existing methods,
and extend the solution to new data ranges ${\mathbb X}$, which have modern
applications.  We assume that the input data is not available, 
precluding machine learning approaches requiring training data pairs $\{\bfx,\bfz\}$.
%Despite this, the results themselves, are very useful in new
%machine learning applications, as we will point out.

\subsection{Maximum Entropy Approaches}
MaxEnt linear feature inversion has been used in
applications including image reconstruction \cite{Gull84,Narayan86,Zhuang87,Zhou,Malik81},
where (\ref{zdef}) represents the point-spread function of the optics,
and in {\it power spectral estimation}, where
 ${\bf W}$ computes the auto-correlation function (ACF) \cite{Kay88}.
%\subsection{Maximum Entropy Approaches}
%The criteria mentioned above lack a principled argument.
%Therefore, we focus on methods that seek to maximize entropy.
There are at least two approaches,
depending on if $\bfx$ is seen as a probability distributon, or as a power 
spectrum \cite{Wernecke77,Narayan86}.  
The MaxEnt principle \cite{Jaynes82,Khinchin} holds that
the data distribution should be selected to maximize uncertainty, measured by 
the distribution entropy
\beq
H_d= - {\cal E}_{\bfx}\left\{ \log p(\bfx) \right\}
=  - \int_{\bf x} \; \log p(\bfx) \; p(\bfx) {\rm d}\bfx.
\label{hddef}
\eeq
While the uniform distribution has the highest entropy 
on a fixed interval, other distributions arise if constraints
are imposed. Constraining variance results in the Gaussian distribution,
and constraining the mean for positive-valued data results in the exponential
distribution \cite{Kapur}.  

%\cite{Csiszar1991},
%\cite{Nadeu1990Flatness},
%\cite{Reitsch77}.
%
%MaxEnt can be seen as a spectral flatness criterion \cite{Nadeu1990Flatness}.
%MaxEnt justification, L2 norm, \cite{Csiszar1991},
%Motovation for MaxEnt and ACG like in Burg \cite{Reitsch77}.
%
%
%\subsection{Applications}
%Applications of deterministic feature inversion include
%data decompression \cite{Luttrell88}, spectral
%estimation \cite{Malik81}, restoration of
%images or data in the event of missing measurements
%\cite{Gull84}, speech reconstruction \cite{milner2002speech},
%feature inversion for time-series
%\cite{Boucheron08,milner2002speech,Chazan},
%image analysis \cite{Vondrick13,dAngelo12,Calonder2010,Alahi2012,
%Oliva06,Weinzaepfel2011},  
%and the visualization of the features used in neural
%networks \cite{Hoggles}.  
%
When $\bfx$ is seen as a probability distribution,
(\ref{hddef}) is applied directly to $\bfx$ 
in the discrete form \cite{Wernecke77,MRF_ref,Zhuang87,Zhou,Gull84}
$ H_{ds}(\bfx) = - \sum_{i=1}^N \;  x_i \; \log x_i,$
where $\sum_{i=1}^N \;  x_i = 1,$
however, is justified 
%This approach is inherently different from (\ref{hddef}),
%because there is no argument that $p(\bfx)$ is
%proportional to $\bfx$ and 
only for strictly positive-valued
$\bfx$. If $\bfx$ is seen as a power spectrum, 
spectral entropy 
\beq
H_s(\bfx) = \sum_{i=1}^N \;  \log x_i
\label{hsdef}
\eeq
is used, where ${\bf W}$ computes the ACF.
Maximizing the spectral entropy maximizes the entropy rate
of a stationary process specified by the
power spectrum $\bfx$ \cite{Malik81,Burg,Burg71,Reitsch77}.
Spectral entropy is also used in image reconstruction
\cite{Wernecke77,Wei87}.

%Below, we will demonstrate that in a special case that the two definitions of
%entropy produce the same result if one is applied to the
%data distribution as a whole, and the other is applied to its mean.
%
Although both distribution entropy and spectral entropy can be seen as 
general smoothness measures \cite{Narayan86,Wernecke77}, a unified
treatment that has place for both would be desireable.
Secondly, as far was we know, prior work on MaxEnt feature inversion  
has been applied to positive-valued data, but not to data
bounded to other ranges, such as the finite interval $[0,\;1]$.

In this paper, we present an approach that can be used
%The solution to MaxEnt linear feature inversion depends
%strongly on the input data range, the prior distribution
%that is used, and constraints that
%one places on the input data moments.  
%This paper presents a solution that can be used for 
with different combinations of data range and constraints.
While classical MaxEnt feature inversion uses just one
combination: positive-valued data with an implicit 
constraint on the mean and exponential prior distribution,
we provide a table listing five combinations, 
all of which are solved be the same universal set of equations.

\section{Idealized Approach}
Our approach consists of two parts, an {\it idealized approach}
which cannot be tractably solved, and an {\it asymptotic approach},
which has a closed-form solution and is asymptotically
the same (as $N$ becomes large) as the idealized approach.

\subsection{Main Idea}
In the idealized approach, we propose a MaxEnt probability
distribution on the set ${\cal M}(\bfz)$,
then let the mean of this distribution
be the MaxEnt solution to the feature inversion problem.
This can be justified because the mean can be seen 
as a smoothing operation applied to an infinite number of samples
from the distribution, desireable from a regularization
argument.  Additionally, notice that ${\cal M}(\bfz)$ 
is the intersection of a linear subspace and $\mathbb{X}$,
which is convex if $\mathbb{X}$ is convex.  The mean of the MaxEnt distribution 
will be the center of gravity on a convex set, 
far from the boundaries,
avoiding artifacts caused when data ``touches" the boundaries.

\subsection{Mathematical Details}
\label{idlsec}
We first assume a MaxEnt prior distribution
on $\mathbb{X}$, written $p_{x,0}(\bfx)$. 
Prior to observing $\bfz$, all we know 
is that $\bfx$ is contained in $\mathbb{X}$.
%It makes sense, therefore, that our knowledge of $\bfx$ takes
%the form of a MaxEnt distribution on $\mathbb{X}$, because this
%expresses the most uncertainty possible.  
%It is not until $\bfz$ has been observed that we can
%constrain our search to for $\bfx$ to ${\cal M}(\bfz)$.
The conditional distribution (given $\bfz$) under the given prior,
is written $p_{x,0}(\bfx|\bfz)$,
has support only on ${\cal M}(\bfz)$ and takes its shape from
$p_{x,0}(\bfx)$. It can be
%Conceptually, $p_{x,0}(\bfx|\bfz)$ is just 
%$p_{x,0}(\bfx)$ constrained to ${\cal M}(\bfz)$
%and normalized, so it integrates to 1.
%and has the s
%The prior still takes effect, so in practical terms, the prior
%$p_{x,0}(\bfx)$ needs to be normalized, so it integrates to 1 
%on ${\cal M}(\bfz)$. This results in a probability distribution
%on ${\cal M}(\bfz)$, written $p(\bfx|\bfz)$. The task, is then to find the mean of this
%conditional distribution, i.e. the conditional mean of $\bfx$ 
%given $\bfz$ under the given prior $p_{x,0}(\bfx)$.  
%Note that ${\cal M}(\bfz)$ has zero volume, since it is an $M$-dimensional 
%subspace in $\mathbb{R}^N$,  and it is limited to the boundaries of
%$\mathbb{X}$.  
represented by 
\beq
p_{x,0}(\bfx|\bfz) = \frac{1}{Z} \delta\left(\bfz-{\bf W}^T \bfx\right) p_{x,0}(\bfx),
\label{mdist}
\eeq
where $\delta$ is either a Dirac delta function,
or an indicator function (equal to 1 if the argument is 0), depending
on how it is used,  and $Z$ is a normalizing factor. 
As an aside, it can be shown \cite{Bag_info} that
$Z=p_{x,0}(\bfz)$, which is $p_{x,0}(\bfx)$ 
mapped to the output of transformation (\ref{zdef})
\footnote{In our notation, the argument of the distribution
defines its region of support, and the subscript 
defines the space where it was defined.
For example, $p_{x,0}(\bfz)$ 
is the image of a distribution defined on
$\mathbb{X}$, but has support on the range of $\bfz$.}.

As a MaxEnt solution, we propose to use the conditional mean of $\bfx$ given $\bfz$
under the given prior $p_{x,0}(\bfx)$
\beq
\bar{\bfx}_z = \int_{\bfx \in {\cal M}(\bfz)} \; \bfx \; p_{x,0}(\bfx|\bfz) \; {\rm d} \bfx,
\label{intxm}
\eeq
which is the weighted center of mass of the
set ${\cal M}(\bfz)$, weighted by $p_{x,0}(\bfx)$. 
Since ${\cal M}(\bfz)$ is bounded by $\mathbb{X}$,
the solution becomes intractable except in the case where  $\mathbb{X}=\mathbb{R}^N$.
%it is difficult to proceed further with a general solution, unless we take specific examples.

\subsection{Data Ranges and Prior Distributions}
We consider three canonical data ranges that are common:
\bitem
\item
$\mathbb{R}^N$ : Unbounded data, ${x_i\in (-\infty , \infty), \forall i}$. 
\item
$\mathbb{P}^N$ : Positive-valued data, ${x_i\in [0 , \infty), \forall i}$. 
\item
$\mathbb{U}^N$ : Doubly bounded data,  ${x_i\in [0  , 1], \forall i}$. 
\eitem
In Table \ref{tab1a},
we list five useful combinations of data range $\mathbb{X}$ and MaxEnt
prior $p_{0,x}(\bfx)$.  Except in $\mathbb{U}^N$, the entropy of a distribution can go to infinity, so it is necessary 
to place constraints on the distribution, either
fixed variance (Gaussian or truncated Gaussian) or
fixed mean (exponential). For the chi-squared distribution with one degree of
freedom, written Chi-sq(1), the mean as well as the mean of $\log x$
is constrained.
\vspace{-.2in}

\begin{table}[!htb]
\caption{MaxEnt priors and activation functions.
TG=``Truncated Gaussian". TED=``Truncated exponential distribution".
${\cal N}\left(x\right) \defined \frac{e^{-x^2/2}}{\sqrt{2\pi}}$ and $\Phi\left( x\right)  \defined \int_{u=-\infty}^x {\cal N}\left(u\right) {\rm d}u.$
%Note that the ``exponential" distribution is more 
%       precicely the chi-squared distribution with 2 degrees of freedom, having mean 2.
}
\begin{center}
\begin{tabular}{|l|l|l|l|l|}
    \hline
$\mathbb{X}$ & \multicolumn{2}{|c|}{MaxEnt Prior} & \multicolumn{2}{|c|}{Activation} \\
    \hline
      &  $p_{0,x}(\bfx)$  &  Name & $\lambda(\alpha)$  & Name \\
    \hline
    $\mathbb{R}^N$   & $\prod_{i=1}^N {\cal N}(x_i)$  & Gaussian & $\alpha$  & linear\\
    \hline
    $\mathbb{P}^N$   & $\prod_{i=1}^N 2 {\cal N}(x_i)$ & Trunc. Gauss. & $\alpha + \frac{{\cal N}(\alpha)}{\Phi(\alpha)}$  & TG \\
    \hline
    $\mathbb{P}^N$   & $\prod_{i=1}^N  e^{-x_i} $ & Expon. & $\frac{1}{1-\alpha}$, $\;\;\alpha<1$& Expon. \\
    \hline
    $\mathbb{P}^N$   & $\prod_{i=1}^N  \frac{e^{-x_i/2}}{\sqrt{2\pi x}} $ & Chi-sq.(1) & $\frac{1}{1-2\alpha}$, $\;\;\alpha<.5$& Ch.Sq(1) \\
    \hline
    $\mathbb{U}^N$  & $\;\;\;\;\;\;$ 1 $\;\;\;\;\;\;$ &   Uniform & $\frac{e^{\alpha}}{e^{\alpha} - 1}-\frac{1}{\alpha}$ & TED \\
    \hline
\end{tabular}
\end{center}
\label{tab1a}
\end{table}
\vspace{-.2in}

\subsection{Unified Exponential form of Prior distribution}
It is well known that MaxEnt priors for moment constraints are
of the exponential class \cite{Kapur}. For
the priors in Table \ref{tab1a},
we can restrict ourselves to the distribution class
consisting of $N$ independent random variables  
\beq
p_s(\bfx;\balpha,\alpha_0,\beta,\gamma)=\prod_{i=1}^N \; p_e(x_i;\alpha_i,\alpha_0,\beta,\gamma),
\label{padef}
\eeq
where $p_e(x;\alpha,\alpha_0,\beta,\gamma)$ is a univariate distribution of the 
following exponential class
\beq
 p_e(x;\alpha,\alpha_0,\beta,\gamma) = \frac{ x^\gamma \; e^{ (\alpha_0+\alpha) x + \beta x^2 }}{ Z(\alpha,\alpha_0,\beta,\gamma)},
\label{expclassdef}
\eeq
where $\balpha=\{\alpha_1, \ldots \alpha_N\}$.
% necessary that $p_e(x;\alpha,\alpha_0,\beta,\gamma)$ integrates to 1.
The MaxEnt prior is then written for $\balpha={\bf 0}$,
\beq
p_{0,x}(\bfx)=p_s(\bfx;{\bf 0},\alpha_0,\beta,\gamma).
\label{p0def}
\eeq
In Table \ref{tab1v} the specific values of $\alpha_0$, $\beta$, and  $\gamma$  are
shown for different data ranges $\mathbb{X}$ and 
moment assumptions, resulting in five different prior distributions.
%Although all parameters are constant, except for
%$\alpha$, we can easily make $\alpha_0$, $\beta$, and  $\gamma$
%also depend on $i$.  This we call non-homogeneous
%data and we will discuss this in Section \ref{nhsec}.

\begin{table}[htb!]
\begin{center}
\caption{Extension of Table \ref{tab1a} showing $\alpha_0$, $\beta$, $\gamma$, and $p_e(x;\alpha,\alpha_0,\beta,\gamma)$.  Key: ``Tr." = Truncated, ``G"=Gaussian, ``Ex."=Exponential}
\begin{tabular}{|c|c|c|c|c|c|}
\hline
$\mathbb{X}$ & $\alpha_0$ & $\beta$ & $\gamma$ & $p_e(x;\alpha,\alpha_0,\beta,\gamma)$  & Name\\
\hline
$\mathbb{R}^N$ & 0 & -.5 & 0 & ${\cal N}(x-\alpha)$ & Gaussian.\\
\hline
$\mathbb{P}^N$ & 0 & -.5 & 0 & $2 {\cal N}(x-\alpha)$ & Tr. G. (TG)\\
\hline
$\mathbb{P}^N$ & -1 & 0 & 0 & $(1-\alpha) e^{-(1-\alpha) x}$ & Expon.\\
\hline
$\mathbb{P}^N$ & -.5 & 0 & -.5 & $\frac{1}{\sqrt{2\pi x(1-2\alpha)}} e^{-\frac{(1-2\alpha)  x}{2}}$ & Chi-Squared(1)\\
\hline
$\mathbb{U}^N$ & 0 & 0 & 0 & $\left(\frac{\alpha}{e^{\alpha} - 1}\right)  \; e^{\alpha x}$ & Tr. Ex. (TED)\\
\hline
\end{tabular}
\label{tab1v}
\end{center}
\end{table}

By comparing Tables \ref{tab1a} and \ref{tab1v},
it can be seen that the MaxEnt priors and activation functions
given in Table \ref{tab1a} are special cases of $p_s(\bfx;\balpha,\alpha_0,\beta,\gamma)$
where $\balpha={\bf 0}$. 
%Later, we'll need to consider non-zero values of $\balpha$.

\section{Asymptotic Approach}
\subsection{Main Idea}
The following closed form approximation to (\ref{intxm}) 
was introduced in \cite{BagUMS}.  
By relaxing the constraint to ${\cal M}(\bfz)$,
we propose a {\it surrogate} distribution with support
everywhere in $\mathbb{X}$,  but having maximum entropy
among all distributions whose mean
is constrained to ${\cal M}(\bfz)$. 
%We call this distribution the {\it surrogate} distribution because
%it is used only as a way to compute the mean.
We later argue that the probability mass of this
surrogate distribution converges to ${\cal M}(\bfz)$,
anyway, and thereby converges (for large $N$) to the idealized $p_{x,0}(\bfx|\bfz)$.

\subsection{Surrogate Distribution}
%In the surrogate distibution approach \cite{BagUMS,BagIcasspPBN},
To approximate  (\ref{mdist}), we
propose form (\ref{padef}),
a proper distribution non-zero everywhere in $\mathbb{X}$.
The surrogate distribution is given by
substituting $\balpha={\bf W} \bfh_z$ into
(\ref{padef}): 
where $\bfh_z$ is value of $\bfh$ that solves
\beq
{\bf W}^T \lambda\left( {\bf W} \bfh\right) = \bfz,
\label{tm1}
\eeq
and $\lambda(\alpha,\alpha_0,\beta,\gamma)$ is the ``activation function",
which calculates the mean of $p_e(x;\alpha,\alpha_0,\beta,\gamma)$.
For simplicity, we drop the dependence 
on $\alpha_0,\beta,\gamma$, which are taken from Table \ref{tab1v}.

We claim that $p_s(\bfx;{\bf W} \bfh_z, \alpha_0,\beta,\gamma)$
approaches (\ref{mdist}) asymptotically as $N$ becomes large,
i.e. the surrogate distribution converges to the posterior $p_{0,x}(\bfx|\bfz)$,
and so the mean of the surrogate distribution, given by
\beq
\bar{\bfx}_z = \lambda({\bf W} \bfh_z),
\label{meanzh}
\eeq
 approaches the mean of $p_{0,x}(\bfx|\bfz)$, and
serves as the MaxEnt solution to the feature inversion problem.
This convergence occurs quickly as a function of $N$ 
(see Section IV.D and Fig. 8 in \cite{BagUMS}).
%We then conclude that as $N$ becomes large, 
%$p_s(\bfx;{\bf W} \bfh_z, \alpha_0,\beta,\gamma) \rightarrow p(\bfx|\bfz)$.
%,
%and (c) has probability mass that concentrates
%on ${\cal M}(\bfz)$. In short, as $N$ becomes large,
%it approximates  (\ref{mdist}).  
%The form of the surrogate distribution depends on $\mathbb{X}$ and $p_{0,x}(\bfx)$ and is given by
%\beq
%p_s(\bfx;\balpha,\alpha_0,\beta)=\prod_{i=1}^N \; p_e(x_i;\alpha_i,\alpha_0,\beta),
%\label{padef}
%\eeq
%where $\balpha=\{\alpha_1, \ldots \alpha_M\}$,
%and $p_e(x;\alpha,\alpha_0,\beta)$ is given in ( \ref{expclassdef}).
%
%We restate the following theorem from \cite{BagUMS,BagIcasspPBN}:
%\begin{theorem}
%Let prior $p_{0,x}(\bfx)$ be written as (\ref{p0def}) , having mean $\lambda({\bf 0})$
%as given in (\ref{l0def}).  Then, the surrogate distribution for (\ref{mdist}) is
%$p_s(\bfx;{\bf W} \bfh_z, \alpha_0,\beta,\gamma)$, where $\bfh_z$ is value of $\bfh$ that solves
%\beq
%{\bf W}^\prime \lambda\left( {\bf W} \bfh\right) = \bfz.
%\label{tm1}
%\eeq
%Interestingly, using the surrogate distribution in place of $p_m(\bfx|\bfz)$
%is the same as using the saddle point approximation to replace
%$p_{0,x}(\bfz)$ in (\ref{jpostdef}) \cite{BagIcasspPBN}.
%\label{thm1}
%\end{theorem}
%For an outline of the proof, see \cite{BagIcasspPBN}.
%It has been shown that the error of the likelihood calculations
%when doing this are negligible \cite{BagPBN,BagSPL2021}.

The surrogate distribution mean $\bar{\bfx}_z$ given by (\ref{meanzh}) enjoys numerous properties.
It can also be shown that $\bfh_z$ is the maximum likelihood estimate of  $\bfh$ under the
likelihood function $p(\bfx;\bfh) = p_s(\bfx;{\bf W}\bfh,\alpha_0,\beta,\gamma)$ \cite{barndorff1979edgeworth}.
As conditional mean estimator, it has well-known optimal properties 
including minimum mean square estimate (MMSE) \cite{KayEst}.
In what follows, we demonstrate the usefulness of (\ref{meanzh})
in linear MaxEnt feature inversion.

Equations (\ref{tm1}) and (\ref{meanzh}),
together with Table \ref{tab1a}, form our unified
approach to MaxEnt feature inversion.
They can be used with a variety of data ranges and 
prior distributions, the most important of which appear in
the table.  Except in the unbounded data case (Gaussian prior),
solving (\ref{tm1}) requires an iterative algorithm
\cite{BagIcasspPBN}.
%We discuss this more in Section \ref{solvsec}.

%n the following, we review examples provided in 
%\cite{BagUMS}, but this time in
%light of the unified solution (\ref{meanzh}), showing the equivalence in each case.

%\subsection{Paper Organization}

\section{Examples}
%We now provide examples from Tables \ref{tab1a} and \ref{tab1v}
%for specific combinations of data range and prior distribution.

\subsection{Unbounded Data}
\label{gausssec}
The unbounded problem $\mathbb{X}=\mathbb{R}^N$ (Gaussian case)
is solved in closed form using the well-known least-squares estimate
\beq
\bar{\bfx}_z= {\bf W} \left({\bf W}^T {\bf W}\right)^{-1} {\bf W}^T \bfx
=  {\bf W} \left({\bf W}^T {\bf W}\right)^{-1} \bfz,
\label{gwcs}
\eeq
which enjoys myriad optimality properties
such as minimum mean square estimator, conditional mean, etc, 
and there is little to add to this.
Equation (\ref{gwcs}) is also a solution to the
idealized formulation (\ref{intxm}).
To see that our unified approach applies,
using Table \ref{tab1a}, we see that $\lambda(\alpha)=\alpha$, 
so the solution $\bfh$ to (\ref{tm1})  is $\left({\bf W}^T {\bf W}\right)^{-1} \bfz$,
and (\ref{gwcs}) follows.

\subsection{Positive-Valued Data}
\label{linexpsec}
The data range $\mathbb{X}=\mathbb{P}^N$,
encompases classical power spectral estimation 
\cite{Malik81,Burg,Burg71,Reitsch77}
and reconstruction of intensity images \cite{Gull84,Narayan86,Zhuang87,Zhou,Malik81}.
%The linear transformation of positive-valued data 
%A solution to the MaxEnt feature inversion for this case
%was presented in \cite{BagUMS}.
%We now show that (\ref{tm1}) and (\ref{meanzh}) 
%do indeed provide the same solution.   
We seek a positive-valued
vector  $\bar{\bfx}$ that meets linear constraint 
\beq
 {\bf W}^T \bar{\bfx} = \bfz. 
\label{lin_constr}
\eeq
The MaxEnt solution presented in \cite{BagUMS} is 
vector  $\bar{\bfx}$, that also meets 
\beq
%\frac{\partial H_e}{\partial u_k} = 
\sum_{i=1}^N\;
\frac{B_{i,k}}{\bar{x}_i} = 0,\;\;\; 1\leq k \leq m,
\label{der0}
\eeq
where $m=N-M$,
and $\{B_{i,k}\}$ is the
%\beq
%\bfx = \bfx_p + {\bf B} {\bf u},
%\label{invt}
%\eeq
%subject to $\bfx\in\mathbb{P}^N$, 
 $N\times(N-M)$ orthonormal matrix
${\bf B}$ spanning the subspace orthogonal to the columns of ${\bf W}$.

This solution results by first seeking a probability distribution with highest entropy
for positive-valued data with given mean $\bar{\bfx}$. This is
the exponential distribution \cite{Kapur},
\beq
    p(\bfx;\bar{\bfx}) \sim
     \prod_{i=1}^N \frac{1}{\bar{x}_i} \; \exp\left\{
	  -\frac{x_i}{ \bar{x}_i}\right\}.
    \label{gy_e}
\eeq
The entropy is then maximized subject to (\ref{lin_constr}).
%with mean parameter $\bar{\bfx} = [\bar{x}_1, \bar{x}_2 \ldots \bar{x}_N]$.
%This is logical, because if we moved outside 
%the column space of ${\bf B}$, then (\ref{lin_constr})
%would no longer hold.
%\it is shown that
%Let $\bfx$ be the length $N$ positive-valued image or
%spectral vector.  
%Let 
%$$\bfz={\bf W}^\prime \bfx,$$ where ${\bf W}$ is $N\times M$
%and $M<N$,
%Examples include autoregressive (AR) analysis,
%which is based on the ACF, MFCC, and DCT analysis of 
%spectra and images.  
%and ${\bf W}$ is full rank.
%
%\subsection{MaxEnt Solution}
%We will derive the MaxEnt solution using fundamental arguments,
%then later show that the solution corresponds to our unified
%solution.
%We note that the distribution with maximum entropy
%for positive-valued data with a specified mean is the exponential
%distribution \cite{Kapur}
%\beq
%    p(\bfx) \sim
%     \prod_{i=1}^N \frac{1}{\bar{x}_i} \; \exp\left\{
%	  -\frac{x_i}{ \bar{x}_i}\right\},
%    \label{gy_e}
%\eeq
%with mean parameter $\bar{\bfx} = [\bar{x}_1, \bar{x}_2 \ldots \bar{x}_N]$.
Note that the entropy of (\ref{gy_e}) is given by \cite{Kapur}
\beq
H_e=\sum_{i=1}^N (1+\log \bar{x}_i).
\label{qme_e}
\eeq
%We propose to maximize the entropy over
%$\bar{\bfx}$ given $\bar{\bfx}\in {\cal M}(\bfz)$,
%i.e. subject to the constraint
%\beq
%{\bf W}^\prime \bfx=\bfz,
%\label{lin_constr}
%\eeq
%where $\mathbb{X}=\mathbb{P}^N$.
%.
As an aside, note that  maximizing (\ref{qme_e})
maximizes the spectral entropy (\ref{hsdef}),
employed in power spectral estimation and sometimes in image reconstruction.

The entropy (\ref{qme_e}) under constraint (\ref{lin_constr})
%\beq
%{\bf W}^\prime \bfx=\bfz,  
%\label{lin_constr}
%\eeq
%where $\mathbb{X}=\mathbb{P}^N$, 
can go to infinity if, for example 
the unity vector ${\bf 1}=[1,1 \ldots 1]$
is orthogonal to the columns of ${\bf W}$.
Then, the mean of $\bfx$ can rise to infinity, together
with the entropy.
To avoid this, we assume that ${\bf 1}={\bf W}{\bf u}$
for some vector ${\bf u}$.
The two equations (\ref{lin_constr}) and (\ref{der0}) provide $N$ equations for 
the $N$ unknowns, but can be solved only if 
there is at least one vector $\bfx\in\mathbb{X}$ meeting (\ref{lin_constr}).

%Solving these equations is discussed in Section \ref{solvsec}.
%
%We wish only to verify that if we apply the unified
%MaxEnt inversion equations (\ref{tm1}) and (\ref{meanzh}),
%that this results in the same solution.
%
%\subsection{Verification of Unified Solution}
We now check if our unified solution works in this case.
By comparing (\ref{gy_e}) with (\ref{expclassdef}),
it can be verified that $\lambda(\;)$
in Table \ref{tab1a} for exponential prior returns the distribution
mean.
%the mean parameter $\bar{x}_i$ is equivalent to
%$\bar{x}_i = -\frac{1}{\alpha_0+\alpha_i}$, which
%is indeed equal to $\lambda(\alpha_i)$.
Now substitute (\ref{meanzh}) into (\ref{tm1}), and get 
${\bf W}^T \bar{\bfx}_z = \bfz$, which meets (\ref{lin_constr}).
Furthermore, note that (\ref{der0}) states that 
the vector $\bar{\bfx}^{-1}=[{1\over \bar{x}_1},
{1\over \bar{x}_2} \ldots {1\over \bar{x}_N}]$
is orthogonal to the columns of ${\bf B}$.
But $\bar{\bfx}^{-1}=-\balpha_0-\balpha,$
where $\balpha_0$ is the constant vector $\balpha_0=\alpha_0 {\bf 1}$.
But $\balpha_0$ is orthogonal to the columns of ${\bf B}$
because we have assumed that ${\bf 1}$
is in the columns space of ${\bf W}$.
Now,  $\balpha$ is the argument of $\lambda(\;)$ in (\ref{tm1}), so 
$\balpha={\bf W}{\bf h}$. Therefore $\balpha$
%Now, since $\balpha$ in (\ref{tm1}) (i.e. the argument of $\lambda(\;)$)
%is in the column space of ${\bf W}$, and 
must also be orthogonal to ${\bf B}$, and (\ref{der0}) is met.
%This validates the use of (\ref{tm1}) and (\ref{meanzh})
%with Tables \ref{tab1a} and \ref{tab1v}.
%\eenum

%Therefore, using (\ref{tm1}) and (\ref{meanzh}) together
%with the exponential prior in Tables \ref{tab1a},
%we solve the special case of MaxEnt linear inversion 
%for positive-valued data. 
%The connection between distribution entropy (\ref{hddef})
%and  spectral entropy (\ref{hsdef}) occurs because, as we 
%see in (\ref{qme_e}), the
%entropy of the exponential distribution
%is indeed a function of the log-spectrum (spectral entropy).
%Therefore, maximizing the distribution entropy of 
%an exponential distribution with a given mean, is the same as
%maximizing the spectral entropy applied to the mean.
%

\subsection{Doubly-Bounded Data}
\label{seclindb}
We now consider data with values to $[0,\;1]$,
such as found in digital photography.
%\subsection{Setup}
%Above, we considered positive data.  We now consider linear transforms
%of data constrained to the range $[0,\;1]$, exemplified by images from digital cameras.
%We adapt the methods of Section \ref{linexpsec}.
The solution has no classical equivalent and results in a novel entropy measure
that was first reported in \cite{BagUMS}, Section V.
The solution parallels the positive data case in the last section
and is based on the truncated exponential distribution
(TED), which is the maximum entropy distribution
for data bounded to the interval $[0,1]$,
under mean (first moment) constraints ( see \cite{Singh2013}, page 186).
The uni-variate TED distribution is provided in Table  \ref{tab1v}
with exponent parameter $\alpha$ that can be positive or negative.

The MaxEnt solution is the vector $\bar{\bfx}$
with elemental values in the range $[0,1]$
that meets (\ref{lin_constr}) and
\beq
\sum_{i=1}^N \; \alpha_i B_{i,k} = 0, \;\;\;, 1\leq k \leq m,
\label{der1um}
\eeq
where the function $\bar{x}=\lambda(\alpha)$
is taken from Table \ref{tab1a} for the TED case.
Since $\lambda(\alpha)$ is strictly monotonically increasing, there is 
a unique $\bar{x}$ for each $\alpha$, so the
$N$-dimensional vectors $\bar{\bfx}$ and $\balpha$
are equivalent parameterizations.
This equation is similar to (\ref{der0}) in the positive data case, and we can make similar
conclusions.  Note that (\ref{der1um}) is equivalent to the statement
that the vector $\balpha$ is in the column space of ${\bf W}$,
or that there exists a free variable ${\bf h}$ such that
$\balpha = {\bf W} {\bf h}$.
It is easy to verify the universal solution
based on (\ref{tm1}) and (\ref{meanzh})
with Tables \ref{tab1a} and \ref{tab1v}.

\subsection{Additional Data Types and Non-Homogeneous Data}
\label{nhsec}
There are cases of interest where
$\bfx$ is non-homogeneous,
having different range $\mathbb{X}$ and 
MaxEnt prior distribution in each of the $N$ elements, 
so that parameters $\alpha_0$, $\beta$, $\gamma$,
and $\mathbb{X}$ in Table \ref{tab1v} depend on the 
element index.  Doing this poses no problem and does not affect the
unified mathematical formulation.
%In Tables \ref{tab1a} and \ref{tab1v},
%we have, in addition to the Gaussian, 
%Exponential, and TED distributions, corresponding
%to data ranges $\mathbb{R}^N$, $\mathbb{P}^N$, and $\mathbb{U}^N$,
In addition to cases we already covered, the
tables provide two distributions for the
positive data case  $\mathbb{P}^N$:
the truncated Gaussian (TG) and Chi-Squared(1)
distributions.  For $\mathbb{P}^N$,
the TG distribution is MaxEnt
for fixed variance and the Chi-Squared(1) distribution is 
MaxEnt when both the mean of $x$ and the mean of $\log(x)$ are
specified (but they cannot be independently specified).

%How do we know which prior distribution to use for $\mathbb{P}^N$
%when there are three possibilities? We look at the way data is generated.
%Consider, for example, a discrete Fourier transform (DFT).
%Due to the central limit theorem, DFT bins tend
%toward either real or complex Gaussian random variables.
%Using the DFT bins directly, real or imaginary parts,
%calls for the unbounded data assumption 
%and the Gaussian prior.
%On the other hand, if a real-valued bin
%is passed through a rectifier, this approximates a truncated Gaussian.
%When computing the  magnitude-square
%of the bins, and assuming a zero mean, 
%the bins is approximated by chi-squared(1)
%for real bins, but by chi-squared(2),
%which is the exponential distribution, for complex bins. 

%\subsection{Discrete Power Spectrum}
%\label{dftsec}
%The most commonly-used non-homogeneous input data
%vector is the magnitude-squared DFT output.
%For even-valued DFT size $N_f$, the DFT will have two real bins ($0$ and $N_f/2$)
%and the remaining $N_f/2-1$ bins are complex.
%Assuming the DFT magnitude-squared has been normalized to have
%a mean value of 1, the magnitude-squared
%of the two real bins will be Chi-sq(1),
%and the complex bins will be exponential.
%The entries in Tables \ref{tab1a} and \ref{tab1v}
%must be made bin-dependent.

%\subsection{Auto-Regressive Power Spectrum}
To demonstrate the use of non-homogeneous data,
%and the connection to classical MaxEnt spectral estimation, 
% we conducted an experiment by 
we passed a Gaussian noise vector of
length $N_f=128$ into a filter,
then computed the DFT, then 
the magnitude-squared of the DFT bins.
This is non-homogeneous because the complex bins produce an exponential distribution,
while the real bins result in Chi-sq(1).
Next, we calculated the ACF
up to an order of $P=6$ using
the classical method: by taking the inverse
DFT. We then set aside the first $N=N_f/2+1$
magnitude-squared DFT bins as ``unknown" input $\bfx$.
Next, we formed the matrix ${\bf W}$
to calculate the inverse DFT from $\bfx$,
resulting in the length $P+1$ ACF.
For this, we set 
$$W_{i,k}=\frac{f(i)}{N_f^2}\cos\left(\frac{2\pi i k}{N_f}\right),$$
for $i=0,1 \ldots N-1$, and $k=0,1 \ldots P$,
where the function $f(i)=2$ when $i=0$ or $i=N-1$
and equals 1 otherwise.  We verified that $\bfz={\bf W}^T \bfx$
indeed equals the ACF up to order $P$.
We then created a bin-dependent function $\lambda(\;)$
implementing the corresponding entries in Table \ref{tab1a}.
We then solved equation (\ref{tm1})
using an off-the shelf optimization program
\footnote{although it is better to use a Newton-Raphson algorithm
that takes advantage of the known Hessian matrix.}
for the variable $\bfh$.
We then used (\ref{meanzh}) to compute $\bar{\bfx}_z$,
the solution to the feature inversion problem.

For comparison purposes, we computed the classical
auto-regressive (AR) spectral estimate
using Levinson algorithm, resulting in the AR spectrum 
$p_{AR}(i) = N \left| \frac{e_0}{A_i}\right|^2,$
where $e_0$ is the AR error variance, and where $\{A_i\}$ are the DFT 
bins of the AR coefficients zero-padded to length $N_f$.
The two spectral estimates, $\bar{\bfx}_z$ and $p_{AR}$,
agreed up to machine precision.
%In Figure \ref{fx_ar_dft}, we see that the
%two spectral estimates, $\bar{\bfx}_z$ and $p_{AR}$
%overlay perfectly.  
%This is expected, because
This one example unifies 
distribution entropy (\ref{qme_e}),
spectral entropy (\ref{hsdef}),
and classical MaxEnt spectral estimation,
using our unified approach
using the nuances of the
non-homogeneous bins.

\subsection{Auto-Encoder}
\label{imgsec}
For a modern example, we implemented feature inversion
of intensity images with pixel values
in $[0,\;1]$, seen as a single-layer auto-encoder
in machine learning terminology. 
Such a situation can also occur in intermediate
network layers, after the use of {\it sigmoid}
activation function and has been demonstrated in
multiple-layer applications \cite{BagEusipcoPBN,BagIcasspPBN}.
Note also that this serves an an kind of optimal auto-encoder
(in the sense of conditional mean) that does not require training,
i.e. the reconstuction network is given, when the analysis network is known.
We used a subset of the MNIST data set consisting of $28\times 28$ images
of handwritten characters.
%We now present 
%to the positive data case in Section \ref{linexpsec}).
%\subsubsection{Cameraman image}
\begin{figure}[!htb]
\begin{center}
\includegraphics[height=0.7in,width=2.9in]{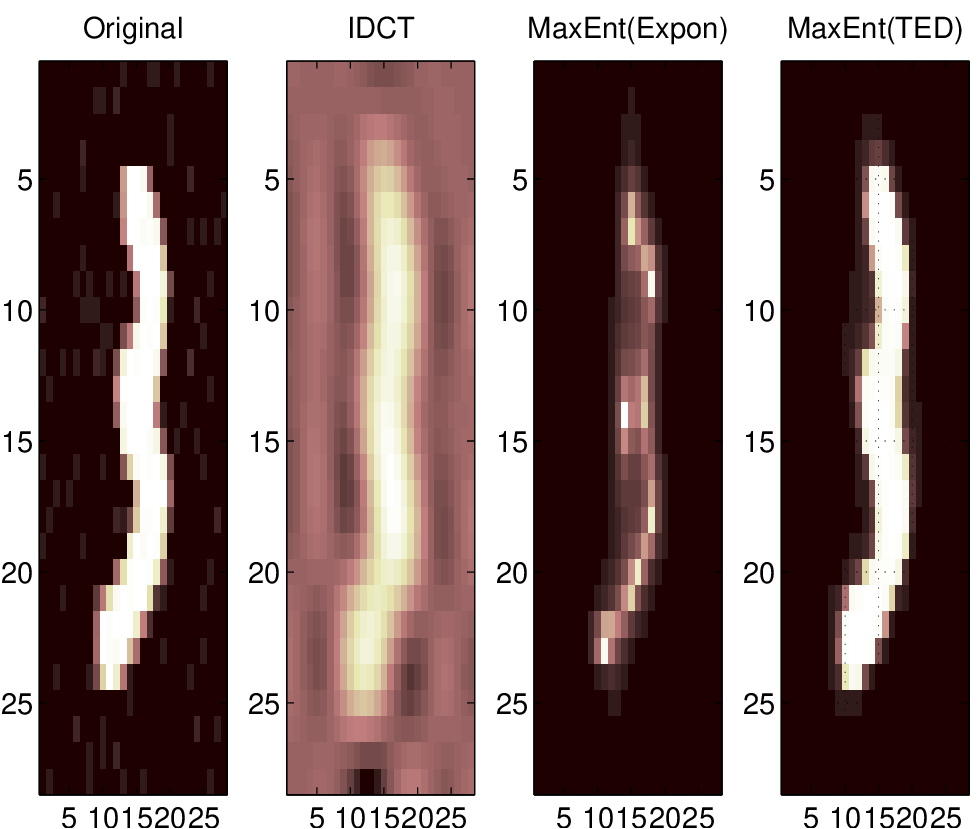}
\includegraphics[height=0.7in,width=2.9in]{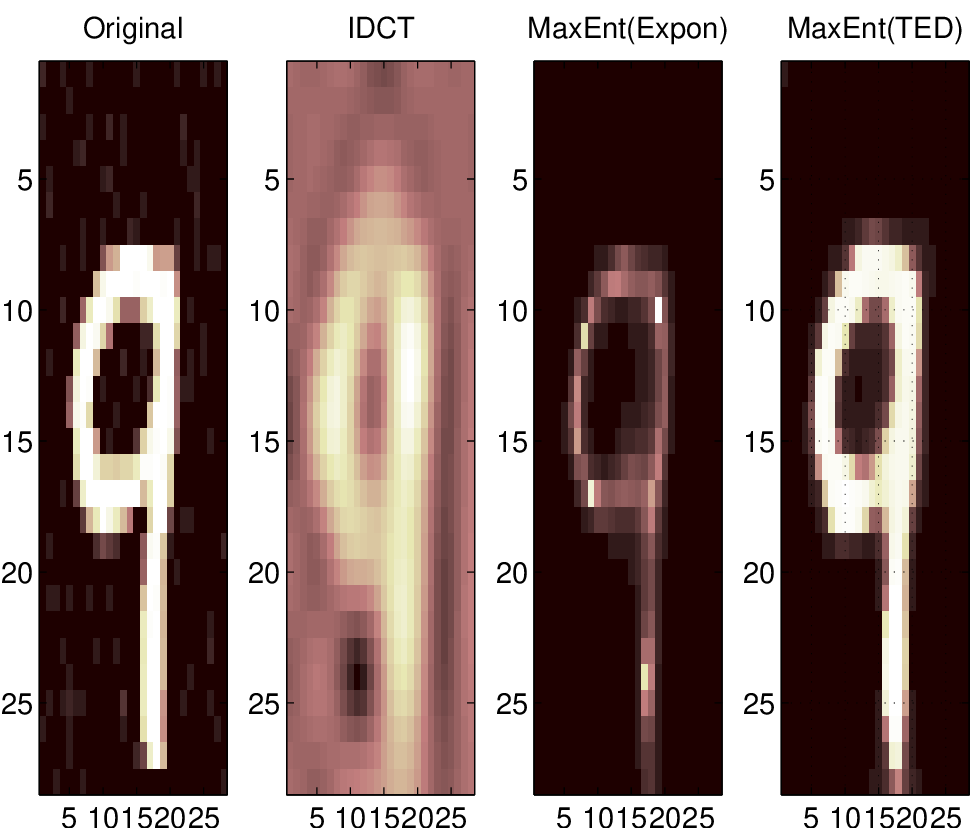}
\includegraphics[height=0.7in,width=2.9in]{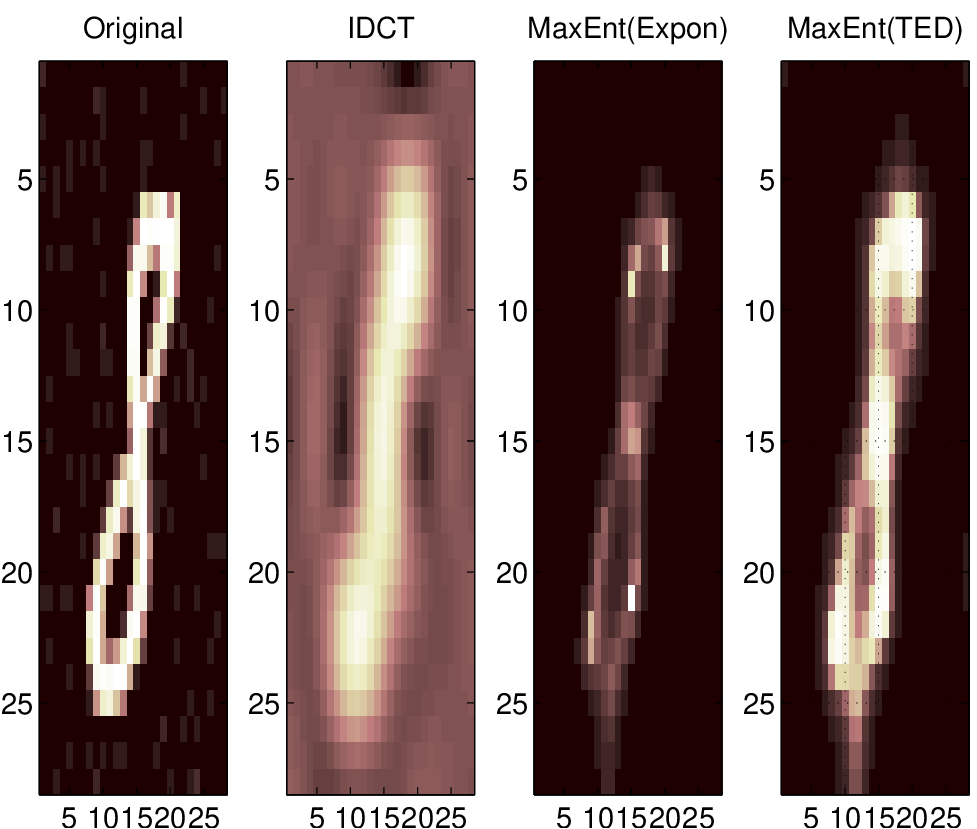}
\includegraphics[height=0.7in,width=2.9in]{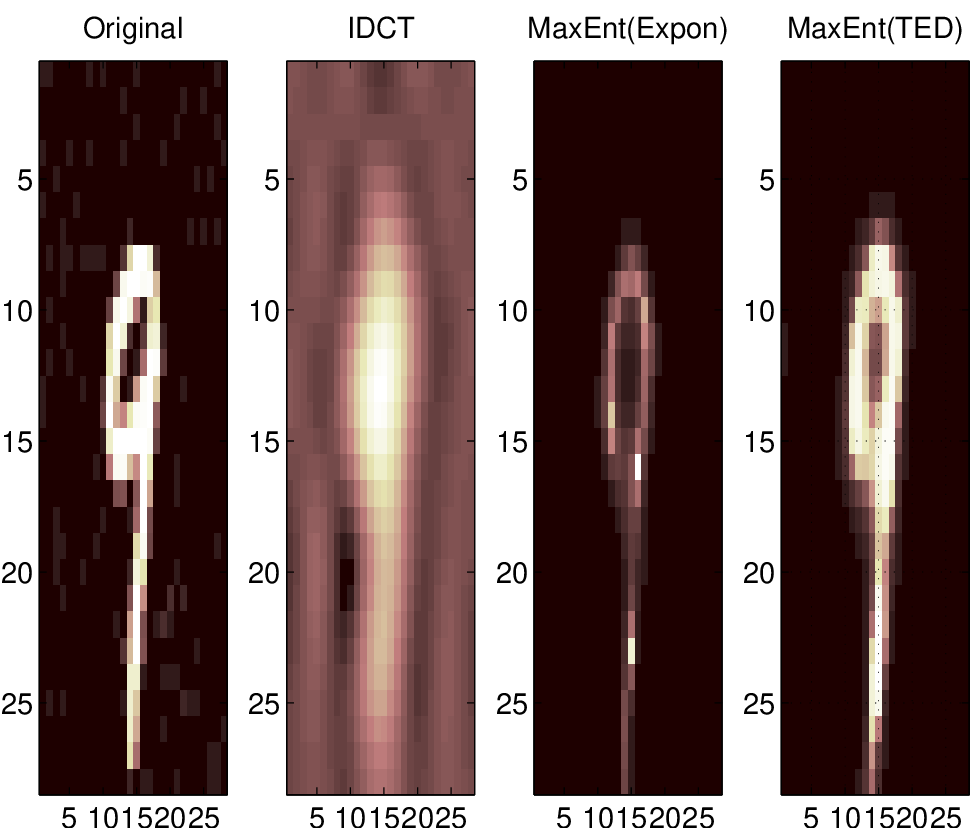}
\includegraphics[height=0.7in,width=2.9in]{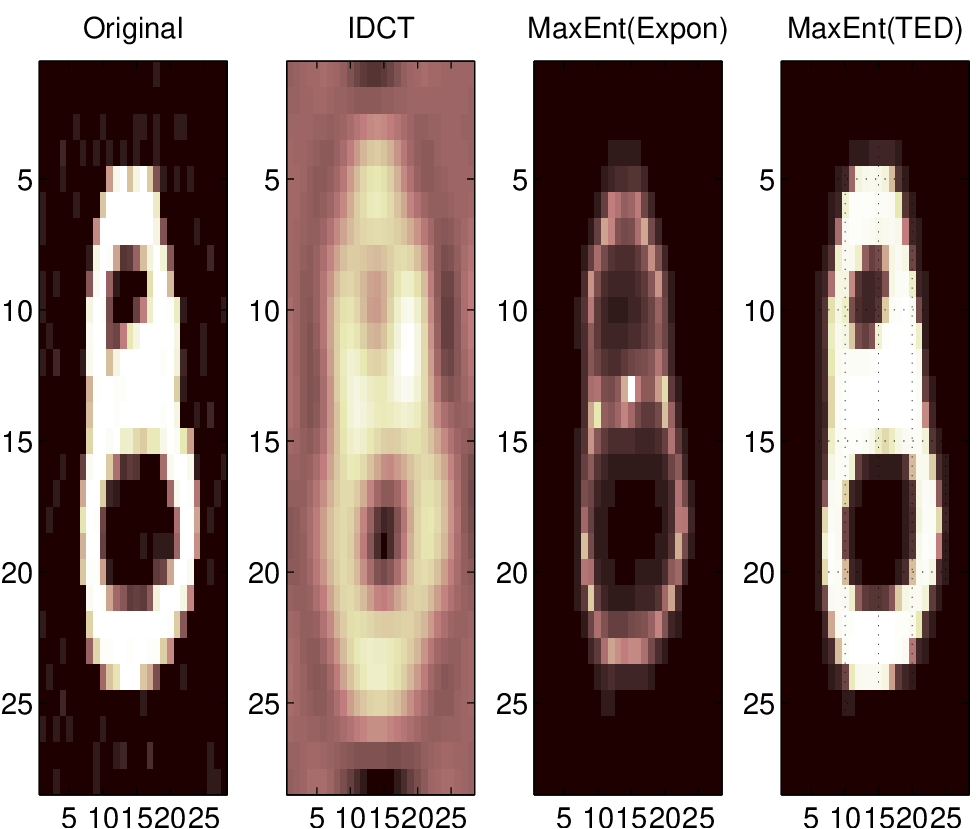}
\caption{$28\times 28$ images reconstructed from $7\times 7$
DCT features. Columns, from left to right: (a) Original,
(b) reconstruction using inverse DCT, 
%same as MaxEnt reconstruction using {\bf unbounded/Gaussian} assumption, 
(c) MaxEnt reconstruction using {\bf positive/Exponential} assumption,
%same as classical MaxEnt image reconstruction
(d) MaxEnt reconstruction using {\bf doubly-bounded/Uniform} assumption.}
\label{cam1}
\end{center}
\end{figure}
Let $\bfx$ be a $n^2\times 1$ vector reshaped from
a square $n\times n$ input image, where $n=28$.
The matrix ${\bf W}$ is constructed to
produce the 2-dimensional discrete cosine transform (DCT)
from the input vector.  In Figure \ref{cam1}, we show six examples.
On the left-most column, the original images are shown.
On the second column, are shown the
reconstructions using inverse DCT from the $7\times 7$ 
DCT coefficients, for a dimension reduction factor of $16$.
These images have negative values
and values greater than 1, so are members of $\mathbb{X}$,
and are very blurry, due to the large reduction in dimension.
On the third column, we show the image reconstructed 
from the same DCT coefficients, using
classical spectral entopy maximization assuming positive data.
This is the formulation used by classical literature in
image reconstruction \cite{Wernecke77}.
There is a very noticeable but artificial improvement in sharpness.
Due to the lack of an upper bound in intensity,
we see a ``glint" effect where some pixel values become large.
Finally, we reconstructed the images using
the doubly-bounded approach (uniform prior).
These images are seen on the right-most column
and look the most like the original images,
preserving the original ``fatness" of the handwritten
characters.

%We show results for a $28\times 28$ image, but for 
% a general n$\times$n image, matrix ${\bf W}$ and the orthogonal
%complement matrix ${\bf B}$ can be huge, $n^2 \times n^2$ taken together.
%Luckily, they do not need to be explicitly constructed. Rather, products
%such as ${\bf W}^\prime \bfx$ or ${\bf B}^\prime \bfx$
%and the derivatives (\ref{der0}) may be computed using the two-dimensional
% DCT and the optimization (\ref{der0}) can be accomplished
%without second derivatives.

\subsection{Conclusions}
In this paper, we have provided a mathematical formulation for MaxEnt
linear feature inversion that handles various data ranges and
MaxEnt prior distributions with a single set of equations. 
We have varified that it works for known examples including
classical image reconstruction and spectral estimation.
In addition, we have provided an example of a new
MaxEnt feature algorithm in an experiment that has
modern applications, where data is constrained to
$[0,1]$.  We demonstrated its superiority over
classical image reconstruction, 
which does not take into account the
upper bound.

\bibliographystyle{ieeetr}
\bibliography{ppt}
\end{document}